\documentclass[runningheads]{llncs}
\usepackage{graphicx}
\usepackage{tikz}
\usepackage{comment}
\usepackage{amsmath,amssymb} % define this before the line numbering.
\usepackage{color}
\usepackage{xcolor}
\usepackage{multirow}
\usepackage{booktabs}
\usepackage[super]{nth}
\usepackage{array}
\usepackage{relsize}
\usepackage{lmodern}
\usepackage{wrapfig}
\usepackage{algorithm,algpseudocode}
\newcommand{\etal}{\textit{et al}. }
\newcommand{\ie}{\textit{i}.\textit{e}.}
\newcommand{\eg}{\textit{e}.\textit{g}.}
\begin{document}
% \linenumbers
\pagestyle{headings}
\mainmatter
\def\ECCVSubNumber{3643}  % Insert your submission number here

\title{Learning Graph Neural Networks for \\Image Style Transfer} % Replace with your title

\begin{comment}
\titlerunning{ECCV-22 submission ID \ECCVSubNumber} 
\authorrunning{ECCV-22 submission ID \ECCVSubNumber} 
\author{Anonymous ECCV submission}
\institute{Paper ID \ECCVSubNumber}
\end{comment}
%******************

% \begin{comment}
\titlerunning{Learning Graph Neural Networks for Image Style Transfer}
\author{Yongcheng Jing\inst{1} \and
Yining Mao\inst{2} \and
Yiding Yang\inst{3} \and
Yibing Zhan\inst{4} \and
Mingli Song\inst{5,2} \and
Xinchao Wang\inst{3} \and
Dacheng Tao\inst{1,4}}
\authorrunning{Y. Jing et al.}
\institute{The University of Sydney, Darlington, NSW 2008, Australia \and
Zhejiang University, Hangzhou, ZJ 310027, China \and
National University of Singapore, Singapore\and
JD Explore Academy, China \and
Zhejiang University City College, Hangzhou, ZJ 310015, China
\\
%\and
%National University of Singapore, Singapore \\
\email{xinchao@nus.edu.sg},
\email{dacheng.tao@gmail.com}}
% \end{comment}
%******************
\maketitle

\begin{abstract}
State-of-the-art parametric and non-parametric style transfer approaches 
{are prone to either distorted} local style patterns due to global statistics alignment, or unpleasing artifacts resulting from patch mismatching.
In this paper, we study a novel semi-parametric neural style transfer framework that alleviates the deficiency of both parametric and non-parametric stylization.   
{The core idea of our approach is to establish accurate and fine-grained content-style correspondences using graph neural networks~(GNNs).
To this end, we develop an elaborated GNN model with content and style local patches as the graph vertices.
The style transfer procedure is then modeled as the attention-based heterogeneous message passing between the style and content nodes in a learnable manner, leading to adaptive many-to-one style-content correlations at the local patch level.}
In addition, an elaborated deformable graph convolutional operation is introduced for cross-scale style-content matching. 
Experimental results demonstrate that the proposed semi-parametric image stylization approach yields encouraging results on the challenging style patterns, preserving both global appearance and exquisite details.
Furthermore, by controlling the number of edges at the inference stage, the proposed method also triggers novel functionalities like diversified patch-based stylization with a single model.
\end{abstract}

\section{Introduction}
\label{sec:intro}

% Background
Image style transfer aims to automatically transfer 
the artistic style from a source style image to a given content one,
and has been studied for a long time in the computer vision community.
Conventionally, image style transfer is generally cast as the problem of non-photorealistic rendering in the domain of computer graphics.
Inspired by the success of deep learning \cite{ding2020self,shen2021training,ding2021understanding,ye2019edge,ding2020context}, Gatys \etal \cite{gatys2016image} pioneer the paradigm that leverages the feature activations from deep \emph{convolutional neural networks (CNNs)} to extract and match the target content and style, leading to the benefits of no explicit restrictions on style types and no requirements of ground-truth training data.
As such, various CNN-based style transfer methods are developed in the literature \cite{kalischek2021light,kolkin2019style,chen2021diverse,xu2021drb,wu2021styleformer,huo2021manifold,hong2021domain,liu2021adaattn,liu2021paint}, establishing a novel field of \emph{neural style transfer (NST)} \cite{jing2019neural}.

    \begin{figure}[!t]
      \centering
      \includegraphics[width=\textwidth]{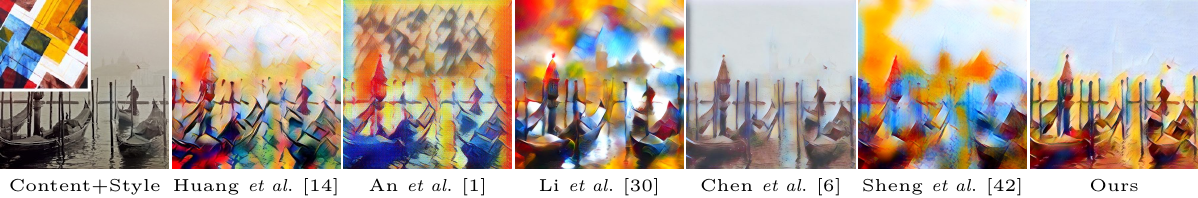}
      % \vspace{-2mm}
      \caption{Existing parametric \cite{huang2017arbitrary,an2021artflow,li2017universal} and non-parametric \cite{chen2016fast,sheng2018neural} NST methods either barely
      transfer the global style appearance to the target \cite{chen2016fast},
      %\xwc{Why they ``fail''? be more specific}
      or produce distorted local style patterns \cite{huang2017arbitrary,an2021artflow,li2017universal} and undesired artifacts \cite{sheng2018neural}.
      By contrast, the proposed GNN-based semi-parametric approach achieves superior stylization performance in the transfers of both global stroke arrangement and local fine-grained patterns.}
      \label{fig:showcase} %% label for entire figure
      % \vspace{-1.5mm}
      \end{figure}

State-of-the-art NST algorithms can be categorized into
two streams of methods,
parametric and non-parametric ones, 
depending on the  style representation mechanisms.
In particular, parametric NST approaches rely on the global summary statistics over the entire feature map from pre-trained deep CNNs to extract and match the target artistic style \cite{gatys2016image,Johnson2016perceptual,huang2017arbitrary}.
Non-parametric neural methods, also known as patch-based NST methods \cite{chen2016fast,sheng2018neural}, leverage the local feature patches to represent the style information, inspired by the conventional patch-based texture modeling approaches with Markov random fields.
The idea is to swap the content neural patches with the most similar style ones, through a greedy one-to-one patch matching strategy.

Both parametric and non-parametric methods, unfortunately, 
have their own limitations, as demonstrated in Fig.~\ref{fig:showcase}.
Parametric stylization methods achieve good performance in transferring the overall appearance of the style images, 
but are incompetent in generating fine-grained local style patterns.
By contrast, non-parametric style transfer algorithms allow for locally-aligned stylization;
however, such patch-based methods are typically accomplished with the undesired artifacts due to content-style mismatching.

In this paper, we present {a semi-parametric style transfer scheme,}
towards alleviating the dilemmas of existing parametric and non-parametric methods.
{On the one hand, our semi-parametric approach allows for the establishment of more accurate many-to-one correspondences between different content and style regions in a learnable manner.}
As such, our approach
explicitly {tackles} the issue of content-style mismatching in non-parametric NST algorithms, thereby 
largely alleviating the deficiency of unplausible artifacts.
{On the other hand, the proposed semi-parametric method adaptively divides content and style features into tiny and cross-scale feature patches for stylization, thus addressing the dilemma of lacking local details in prior parametric schemes.
}

Towards this end, we introduce to the proposed 
semi-parametric NST a dedicated learning mechanism, 
\emph{graph neural networks}~(GNNs), to enable  
adaptive local patch-level interplay between the content and style.
As a well-established learning paradigm
for handling non-Euclidean data,
GNNs are designed to explicitly account for  
structural relations and interdependency between
nodes. Moreover, GNNs are equipped with
efficacious strategies for aggregating information
from multiple neighbors to a center node. 
Such competences make GNN
an ideal tool for tackling the
intricate  content-style region matching challenge
in style transfer, especially the 
many-to-one mapping between each content
patch and multiple potentially-matching style patches.
We therefore exploit GNNs to adaptively set 
up the faithful topological correspondences
among the very different content and style, 
such that every content region is rendered 
with the optimal style strokes.

Specifically, we start by building a heterogeneous NST graph, with content and style feature patches as the vertices.
The multi-patch parametric aggregation in semi-parametric NST can thereby be modeled as the message passing procedure among different patch nodes in the constructed stylization graph.
By employing the prevalent GNN mechanisms such as the graph attention network, the $k$ most similar patches can be aggregated in an attention-based parametric manner. 
The aggregated patches are then composed back into the image features, which are further aligned with the target global statistics to obtain the final stylized result.  
Also, a deformable graph convolutional operation is devised, making it possible for cross-scale style-content matching with spatially-varying stroke sizes in a single stylized image.
Furthermore, our GNN-based NST can readily perform diversified patch-based stylization, by simply changing the number of connections during inference.

{In sum, our contribution is a novel semi-parametric arbitrary stylization 
scheme that allows for the effective generation of both 
the global and local style patterns, backed by a dedicated deformable graph convolutional design.}
This is specifically achieved through modeling the NST process 
as the message passing between content and style under the framework of GNNs.
Experimental results demonstrate that the proposed GNN-based stylization method yields results superior to the state of the art.

\section{Related Work}
\label{sec:related_work}

\noindent\textbf{Neural Style Transfer.} 
Driven by the power of \emph{convolutional neural networks (CNNs)} \cite{zhan2020multi,zhao2020collaborative,zhan2019exploring,kong2021adaptive}, Gatys \etal propose to leverage CNNs to capture and recombine the content of a given photo and the style of an artwork \cite{gatys2016image}, leading to the area of \emph{neural style transfer (NST)}.
Existing NST approaches can be broadly divided into parametric and non-parametric NST methods. 
Specifically, parametric NST approaches leverage the global representations to transfer the target artistic style, which are obtained by computing the summary statistics in either an image-optimization-based online manner \cite{gatys2016image,li2017demystifying,wilmot2017stable,liu2021learning}, or model-optimization-based offline manner \cite{Johnson2016perceptual,zhang2017multi,li2017diverse,wang2021rethinking,an2021artflow,liu2021adaattn,chen2017stylebank,li2017universal,huang2017arbitrary,chen2020explicit,jing2020dynamic}.
On the other hand, non-parametric methods exploit the local feature patches to represent the image style \cite{li2016combining,sheng2018neural,champandard2016semantic,chen2016fast,mechrez2018contextual,liao2017visual}, inspired by the conventional patch-based texture modeling approaches with Markov random fields.
The idea is to search the most similar neural patches from the style image that match the semantic local structure of the content one \cite{li2016combining,sheng2018neural,champandard2016semantic,chen2016fast,mechrez2018contextual,liao2017visual}.
This work aims to seek a balance between  parametric and non-parametric NST methods by incorporating the use of GNNs.

\noindent\textbf{Graph Neural Networks.}
GNNs have merged as a powerful tool to handle graph data in the non-Euclidean domain~\cite{kipf2017semi,velivckovic2018graph,jing2021amalgamating,jing2021meta,yang2020factorizable,yang2021spagan,yang2020distilling,yang2021learning,Huihui21AAAI}.
In particular, the seminal work of Kipf and Welling \cite{kipf2017semi} proposes graph convolutional networks (GCNs), which successfully generalizes CNNs to deal with graph-structured data, by utilizing neighborhood aggregation functions to recursively capture high-level features from both the node and its neighbors.
The research on GNNs leads to increasing interest in deploying GNN models in various graph-based tasks, where the input data can be naturally represented as graphs \cite{zhou2018graph}.
Moreover, the emerging transformers can also be treated
as generalizations of GNNs~\cite{Sucheng2022CVPR,Weihao22MetaFormer,zhang2022vsa,zhang2022vitaev2,xu2021vitae}.
Unlike these existing works where the inputs are themselves non-grid graphs, we aim to extend the use of GNN models to effectively manipulate grid-structured images, such that various image-based tasks can be benefited from GNNs.

\section{Proposed Method}

Towards addressing the limitations of existing parametric and non-parametric NST methods, we introduce the proposed semi-parametric style transfer framework with GNNs.
In what follows, we begin by providing an overview of the proposed GNN-based approach, and then elaborating several key components, including the construction of the topological NST graph, the dedicated deformable graph convolutional operation customized for the established NST graph, and the detailed 2-hop heterogeneous message passing process for stylization.
Finally, we illustrate the cascaded patch-to-image training pipeline, tailored for the proposed GNN-based stylization system.

\begin{figure}[!t]
  \centering
  \includegraphics[width=0.9\textwidth]{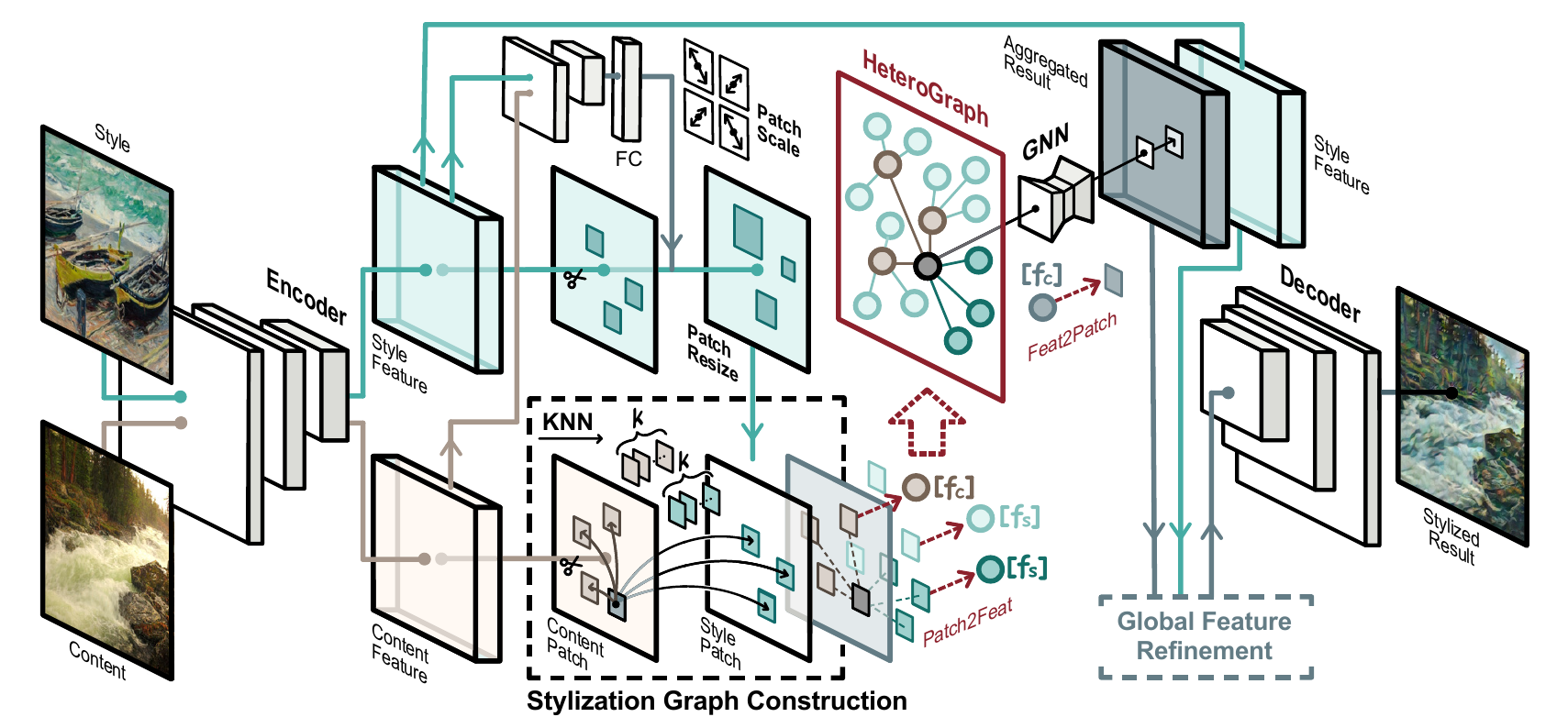}
  % \vspace{-2mm}
  \caption{Network architecture of the proposed semi-parametric style transfer network with GNNs. 
  From left to right, the corresponding stylization pipeline comprises four subprocesses, \ie, image encoding with the encoder, local patch-based manipulation based on heterogeneous GNNs, global feature refinement, and the feature decoding procedure. 
  The symbols of scissors represent the process to divide the feature maps into  feature patches.
  HeteroGraph denotes the established heterogeneous stylization graph with two types of content-style inter-domain connections and content-content intra-domain connections.}\label{fig:arch}
\end{figure}

\subsection{Network Overview}

The overall workflow of the proposed semi-parametric NST framework is shown in Fig.~\ref{fig:arch}.
There are primarily four modules in the whole pipeline, termed as \emph{image encoding}, \emph{local patch-based manipulation}, \emph{global feature refinement}, and \emph{feature decoding}. 
At the heart of the proposed framework is the \emph{local patch-based manipulation} module, which will be further detailed in the following sections.

% \vspace{1mm}
\noindent\textbf{Image Encoding Module.} The proposed semi-parametric stylization starts by receiving style and content images as inputs and encoding these images into meaningful feature maps (the green and yellow blocks in Fig.~\ref{fig:arch}), by exploiting the first few layers of the pre-trained VGG network.
In particular, unlike the existing work \cite{huang2017arbitrary} that uses the layers before $\texttt{relu4\_1}$, we leverage the VGG layers up to $\texttt{relu3\_1}$, for the sake of more valid feature patches that can be exploited by the following local patch-based feature transformation stage.  

% \vspace{1mm}
\noindent\textbf{Local Patch-based Manipulation Module.} With the embeded content and style features as inputs, the local patch-based manipulation module extracts the corresponding content and style feature patches with the stride of $s$ and the sliding window size of $p \times p$, represented as the scissor symbol in Fig.~\ref{fig:arch}. 
We then build a heterogeneous stylization graph (the red frame in Fig.~\ref{fig:arch}) with the obtained feature patches as graph nodes and perform the dedicated deformable graph convolution to generate the locally-stylized features, which will be further detailed in the succeeding Sect.~\ref{sect:graph_construction} and Sect.~\ref{sect:deformable}.

% \vspace{1mm}
\noindent\textbf{Global Feature Refinement Module.} The produced style-transferred results from the stage of patch-based manipulation are effective at preserving fine-grained local style patterns; however, the global style appearance is likely to be less similar to the target style image, due to the lack of global constraint on the stroke arrangement.
To alleviate this dilemma, we propose a hierarchical patch-to-image stylization scheme to yield both the exquisite brushstroke and large-scale texture patterns.
This is achieved by refining the feature representations at a global level, subsequent to the local patch-based manipulation.
For the specific refinement method, since there already exist several effective global feature decorated strategies in the field of NST (\eg, adaptive instance normalization (AdaIN) \cite{huang2017arbitrary} and zero-phase component analysis (ZCA) \cite{li2017universal}), here we directly utilize AdaIN as our refinement scheme, considering its high efficiency.

% \vspace{1mm}
\noindent\textbf{Feature Decoding Module.} The last stage of our semi-parametric style transfer pipeline, termed as feature decoding, aims to decode the obtained feature representations from the preceding global feature refinement module into the final stylized image.
The decoder module specifically comprises a sequence of convolutional and bilinear upsampling layers with the ReLU nonlinearities.

% \vspace{1mm}
In the following sections, we will explain more details regarding the key module of \emph{Local Patch-based Manipulation} with GNNs, including the graph construction procedure and the deformable graph convolutional process.

\subsection{Stylization Graph Construction}
\label{sect:graph_construction}

At the stage of local patch-based manipulation, the first challenge towards the adaptive patch-level interactions between content and style with GNNs is the establishment of topological graphs. 
Unlike conventional GNN-based applications where the inputs can be naturally modeled as graphs (\eg, biological molecules and social networks), 
there is no such natural topological structure for our task of semi-parametric image style transfer.
To address this issue, we develop a dedicated graph construction technique, tailored for image stylization.

We start by giving the mathematical model of general graph-structured data as: $\mathcal{G}=\{\mathcal{V}, \mathcal{E}\}$, where $\mathcal{G}$ represents a directed or undirected graph.
$\mathcal{V}$ denotes the set of vertices with nodes $v_i \in \mathcal{V}$.
$\mathcal{E}$ represents the edge set with $(v_i, v_j) \in \mathcal{E}$, where $\{v_j\}$ is the set of neighboring nodes of $v_i$.
Each vertex has an associated node feature $\mathcal{X} = [x_1 \ x_2 \ ... \ x_n]$.
For example, $x$ can be defined as the 3D coordinates in the task of point cloud classification.

As can be observed from the above formulation of prevalent graph data, the key elements in a graph are the vertices with the corresponding node features as well as the edges, which are thereby identified as our target objects to instantiate in the domain of style transfer as follows:

% \vspace{1mm}
\noindent\textbf{Heterogeneous Patch Vertices.} 
To leverage GNNs to benefit the local-level stylization, we model in our framework the content and style patches as the graph nodes.
Specifically, we exploit the content and style feature activations from the pre-trained VGG encoder, shown as the green and yellow blocks in Fig.~\ref{fig:arch}, respectively, to capture the corresponding feature patches with a sliding window (\ie, the scissor symbol in Fig.~\ref{fig:arch}), in a similar manner as what is done when performing convolutions. 
We set the stride as 1 by default, meaning that there exist overlaps among our extracted activation patches.
Such a manner of overlapped patch generation allows for smooth transitions among different stylized regions. 
In particular, to achieve cross-scale patch matching, we perform multi-scale patch division, which will be demonstrated in detail as a part of the deformable convolution in Sect.~\ref{sect:deformable}.

For the definition of the associated features for each patch vertex, we use a $\texttt{Patch2Feat}$ operation, depicted as the red fonts in Fig.~\ref{fig:arch}, to produce the desired format of node features for the use of the subsequent GNN layers, as also done in \cite{zhou2020cross}.
The designed $\texttt{Patch2Feat}$ operation specifically amalgamates the $c$-dimensional features at each position of the $p \times p$ activation patch into a 1-dimensional feature vector, which is then considered as the node feature at every patch vertex.
The derived content and style node features are shown as $[f_c]$ and $[f_s]$ in Fig.~\ref{fig:arch}, respectively, for the use of the latter GNN layers.

% \vspace{1mm}
\noindent\textbf{Inter- and Intra-KNN Edges.} 
Another critical issue in building the stylization graph is the establishment of connections among different patch vertices.
Customized for the task of style transfer, we formulate two types of edges, termed as \emph{content-style inter-domain edges} and \emph{content-content intra-domain edges}, leading to a special kind of heterogeneous graph.

In particular, the inter-domain connections between heterogeneous style and content nodes aim to attain more accurate many-to-one style-content matching for patch-based stylization.
More specifically, for each content query patch $\phi_{i}(\mathcal{F}_c)$ with $\mathcal{F}_c$ representing the whole content feature map, we search the corresponding $k$-nearest ones in the set of style feature patches $\{\phi(\mathcal{F}_s)\}$, which are identified as the neighbors coupled with inter-domain edges.
This process of $k$-nearest neighbor search (KNN) is shown in the black dotted frame in Fig.~\ref{fig:arch}.
We employ the distance metric of normalized cross-correlation (NCC) for pair-wise KNN, by scoring the cosine distance between a couple of content and style patches.
Given a specific content patch $\phi_i(\mathcal{F}_c)$ as the query, 
our KNN procedure based on NCC can be specifically formulated as:
\begin{equation}
  % \begin{aligned}
  \mathrm{KNN}(\phi_i(\mathcal{F}_c), \{\phi(\mathcal{F}_s)\}) = 
  \underset{j\in\{1,\ldots,N_s\}}{\arg{\max}_\mathrm{k}} \frac{\langle\phi_i(\mathcal{F}_c), \phi_j(\mathcal{F}_s)\rangle}{\|\phi_i(\mathcal{F}_c)\| \|\phi_j(\mathcal{F}_s)\| }, i\in\{1,\ldots,N_c\},
% \end{aligned}
\label{eq:knn}
\end{equation}
where $N_c$ and $N_s$ denote the cardinalities of the corresponding content and style patch sets, respectively. 
$\max_\mathrm{k}$ returns the $k$ largest elements from the set of the computed pair-wise NCCs.
$\mathrm{KNN}(\phi_i(\mathcal{F}_c))$ represents the target $k$ nearest-neighboring style vertices for the content patch $\phi_i(\mathcal{F}_c)$.

We also introduce the intra-domain connections within the set of content activation patches in our stylization graph, shown as the brown arrows in the black dotted rectangle in Fig.~\ref{fig:arch}.
The goal of such content-to-content edges is to unify the transferred styles across different content patches.
In other words, we utilize the devised intra-domain connections to make sure that the semantically-similar content regions will also be rendered with homogeneous style patterns.  
This is specifically accomplished by linking the query content patch $\phi_i(\mathcal{F}_c)$ with the top-$k$ most similar patches $\{\phi_j(\mathcal{F}_c)\}$ where $j\in\{1,\ldots,N_c\}$, by NCC-based KNN search in a similar manner with that in Eq.~\ref{eq:knn}.

% \vspace{1mm} 
The ultimate heterogeneous stylization graph, with the two node types of content and style vertices and also the two edge types of inter- and intra-domain connections, is demonstrated as the red rectangle in Fig.~\ref{fig:arch}.
The relationship between the involved nodes is defined as the NCC-based patch similarity.

\subsection{Deformable Graph Convolution}
\label{sect:deformable}

With the constructed stylization graph, we are then ready to apply GNN layers to perform heterogeneous message passing along the content-style inter-domain edges and also content-content intra-domain edges.
A na\"ive way will be simply performing existing graph convolutions on the heterogeneous stylization graph to aggregate messages from the content and style vertices.

However, this vanilla approach is not optimal for the task of style transfer, due to a lack of considerations in feature scales.
Specifically, in the process of image stylization, the proper feature scale is directly correlated with the stroke scale in the eventual output \cite{jing2018stroke}, which is a vital geometric primitive to characterize an artwork.
The objective stylized results should have various scales of style strokes across the whole image, depending on the semantics of different content regions.

Towards this end, we propose a dedicated deformable graph convolutional network that explicitly accounts for the scale information in message passing.
The devised deformable graph convolutional network comprises two components.
Specifically, the first component is an elaborated \emph{deformable scale prediction module}, with a fully-connected (FC) layer in the end, that aims to generate the optimal scale of each patch in a learnable manner before conducting message aggregation, as also done in \cite{chen2021dpt}.
In particular, the scale predictor receives both the content and style features as inputs, 
considering the potential scale mismatching between the content and style, as shown in the upper left part of Fig.~\ref{fig:arch}.

As such, by adaptively performing scale adjustment according to both content and style inputs,
the proposed deformable graph conventional network makes it possible for cross-scale style-content matching with spatially-varying stroke sizes across the whole image.
We clarify that we only incorporate one-single predictor in our deformable graph convolutional network that produces the style scales, for the sake of computational efficiency.
There is no need to also augment another predictor for content scale prediction, which is, in fact, equivalent to fixing the content scale and only changing the style one.

The second component of the proposed deformable graph convolutional network is the \emph{general feature aggregation module} that learns to aggregate the useful features from the neighboring heterogeneous content and style nodes.
Various existing message passing mechanisms can, in fact, readily be applied at this stage for message propagation.
Here, we leverage the graph attention scheme to demonstrate the message flow along with the two types of stylization edges, which empirically leads to superior stylization performance thanks to its property of anisotropy.

Specifically, given an established stylization graph, our dedicated heterogeneous aggregation process is composed of two key stages, termed as \emph{style-to-content message passing stage} and \emph{content-to-content messing passing stage}:

% \vspace{1mm}
\noindent
\textbf{Style-to-Content Message Passing.} The first style-to-content stage aims to gather the useful style features from the $k$ neighboring style vertices.
For the specific message gathering method, one vanilla way is to treat the information from every style vertex equally, meaning that the aggregated result would be simply the sum of all the neighboring style node features.
However, the results of such na\"ive approach are likely to be affected by the noisy style vertices, resulting in undesired artifacts.

To tackle this challenge, we apply an attention coefficient for each style vertex during message passing, which is learned in a data-driven manner.
Given a centering content node $v_c$ and its neighboring style nodes $\{v_s\}$ with the cardinality of $k$, the learned attention coefficients $w(v_c,v_s^j)$ between $v_c$ and a specific neighbor $v_s^j$ can be computed as:
\begin{equation}
  % \begin{aligned}
  w(v_c,v_s^j) =  
  \frac{\exp\left(\text{LeakyReLU}\left({ W_a}[{ W_b}\mathcal{F}_c\|{ W_b}\mathcal{F}_s^j]\right)\right)}{\sum_{m=1}^k \exp\left(\text{LeakyReLU}\left({ W_a}[{ W_b}\mathcal{F}_c\|{ W_b}\mathcal{F}_s^m]\right)\right)},
  % \end{aligned}
  \label{eq:atten}
\end{equation}
where $W$ represents the learnable matrix in linear transformation.  
$\|$ is the concatenation operation.

With such an attention-based aggregation manner, our stylization GNN can adaptively collect more significant information from the best-matching style patches, and meanwhile reduce the features from the less-matching noisy ones.
Furthermore, we also apply a multi-headed architecture that generates the multi-head attention, so as to stabilize the attention learning process.

% \vspace{1mm}
\noindent
\textbf{Content-to-Content Message Passing.} 
With the updated node features at the content vertices from the preceding style-to-content message passing process, we also perform a second-phase information propagation among different content nodes.
The rationale behind our content-to-content message passing is to perform global patch-based adjustment upon the results of the style-to-content stage, by considering the inter-relationship between the stylized patches at different locations.
As such, the global coherence can be maintained, where the content objects that share similar semantics are more likely to resemble each other in stylization, which will be further validated in the experiments.

This proposed intra-content propagation also delivers the benefit of alleviating the artifacts resulting from potential style-content patch mismatching, by combining the features from the correctly-matching results.
The detailed content-to-content message passing procedure is analogous to that in style-to-content message passing, but replacing the style vertices in Eq.~\ref{eq:atten} with the neighboring content vertices with the associated updated node features.

% \vspace{1mm}
The eventual aggregation results from the proposed inter- and intra-domain message passing are then converted back into the feature patches by a $\texttt{Feat2Patch}$ operation, which is an inverse operation of $\texttt{Patch2Feat}$.
The obtained patches are further transformed into the feature map for the use of the subsequent global feature alignment module and feature decoding module.

\subsection{Loss Function and Training Strategy}

To align the semantic content, our content loss $\mathcal{L}_{c}$ is defined as the perceptual loss over the features from layer $\{\texttt{relu4\_1\}}$ of the pre-trained VGG network $\Phi$:
\begin{align}
\begin{split}
\mathcal{L}_{c}
= \| \Phi^{\texttt{relu4\_1}} ( \mathcal{I}_{c} )-\Phi^{\texttt{relu4\_1}} ( \mathcal{I}_{o}  )  \|_{2},
\end{split}
\label{eq:closs}
\end{align}
where $\mathcal{I}_{c}$ and $\mathcal{I}_{o}$ represent the content and the output stylized images, respectively.
For the style loss, we use the BN-statistic loss to extrat and transfer the style information, computed at layer $\{ \texttt{relu1\_1}, \texttt{relu2\_1}, \texttt{relu3\_1}, \texttt{relu4\_1} \}$ of the VGG network $\Phi$:
\begin{align}
% \vspace{-1mm}
\begin{split}
\mathcal{L}_{s}(h)&= \sum_{\ell=1}^{4}
% \left( 
\mathlarger{(}
\left \| h \left( \Phi^{\texttt{relu}\ell\texttt{\_1}} \left ( \mathcal{I}_{s} \right) \right)- h \left ( \Phi^{\texttt{relu}\ell\texttt{\_1}} \left( \mathcal{I}_{o} \right) \right) \right \|_{2} 
\mathlarger{)},
% \right) 
\end{split}
% \vspace{-1mm}
\label{eq:sloss}
\end{align}
where $h(\cdot)$ denotes the mapping of computing the BN statistics over the feature maps.
The style loss can then be defined as:
$\mathcal{L}_s=\mathcal{L}_s(\mu)+\mathcal{L}_s(\sigma)$, with $\mu(\cdot)$ and $\sigma(\cdot)$ denoting mean and standard standard deviation, respectively.

Our total loss is thereby a weighted sum of the content and style loss, formulated as: $\mathcal{L} = \mathcal{L}_{content} + \lambda \mathcal{L}_{style}$ with $\lambda$ as the weighting factor that balances the content and style portions.

{
    \setlength{\textfloatsep}{2pt}
    \renewcommand{\algorithmicrequire}{\textbf{Input:}} 
    \renewcommand{\algorithmicensure}{\textbf{Output:}}
    \begin{algorithm}[!t]
    % \small
      \begin{algorithmic}[1]
              \Require{
                $\mathcal{I}_c$: the content image; $\mathcal{I}_s$: the style image; $\texttt{VGG}$: the pre-trained loss network.
                }
        \Ensure{$\mathcal{I}_o$: Target stylized image that simultaneously preserves the appearance of $\mathcal{I}_s$ and the semantics of $\mathcal{I}_c$.}
        % -------------------------------------------------------------
        \State Perform initializations on the image encoder $\texttt{Enc}(\cdot)$, the scale predictor $\texttt{Prec}(\cdot)$, GNN parameters $W_a$ and $W_b$, and the feature decoder $\texttt{Dec}(\cdot)$. 
        \For{$i=1$ to $\mathcal{T}$ iterations}
        \State Feed $\mathcal{I}_s$ and $\mathcal{I}_c$ into $\texttt{Enc}(\cdot)$ and obtain the style and content features 
        $\mathcal{F}_s$ and $\mathcal{F}_c$;
        \State Divide $\mathcal{F}_c$ into equal-size content patches $\{\phi(\mathcal{F}_c)\}$
         by using a sliding window;% with the stride $s$;
        \State Feed \{$\mathcal{F}_s$, $\mathcal{F}_c$\} into $\texttt{Prec}(\cdot)$ and obtain the optimal scales $\{\alpha\}$ for style patches;
        \State Divide $\mathcal{F}_s$ into varying-size style patches $\{\phi(\mathcal{F}_s)\}$ with the obtained scales $\{\alpha\}$;
        \State Resize $\{\phi(\mathcal{F}_s)\}$ according to the size of the content patches $\{\phi(\mathcal{F}_c)\}$;
        \State Construct inter- and intra-domain edges by Eq.~\ref{eq:knn};
        \State Transform $\{\phi(\mathcal{F}_s)\}$ and $\{\phi(\mathcal{F}_c)\}$ into the node features by using $\texttt{Patch2Feat}$;
        \State Establish the heterogeneous graph $\mathcal{G}_{NST}$ and feed $\mathcal{G}_{NST}$ into the GNN layers;
        \State Perform heterogeneous message passing over $\mathcal{G}_{NST}$ by Eq.~\ref{eq:atten} and obtain $\texttt{f}_\texttt{c}$;
        % obtain the aggregated results $\texttt{f}_\texttt{c}$;
        \State Convert the aggregation results $\texttt{f}_\texttt{c}$ into feature map $\mathcal{F}_o$ by $\texttt{Feat2Patch}$;
        \State Feed the obtained features $\mathcal{F}_o$ into the global feature refiner and obtain $\mathcal{F}_o'$;
        \State Feed $\mathcal{F}_o'$ into the decoder $\texttt{Dec}(\cdot)$ to obtain the target stylized image $\mathcal{I}_o$;
        \State Feed \{$\mathcal{I}_o$, $\mathcal{I}_c$, $\mathcal{I}_s$\} into $\texttt{VGG}$ and compute $\mathcal{L}_{c}$ and $\mathcal{L}_{s}$ by Eq.~\ref{eq:closs} and Eq.~\ref{eq:sloss};
        \State Optimize $\texttt{Enc}(\cdot)$, $\texttt{Prec}(\cdot)$, $W_a$, $W_b$, and $\texttt{Dec}(\cdot)$ with the Adam optimizer;
        \EndFor
      \end{algorithmic}
        \caption{Training a GNN-based stylization model that can transfer arbitrary styles in a semi-parametric manner.}
        \label{alg:A}
        
    \end{algorithm}
}

We also derive an elaborated training pipeline, tailored for the proposed GNN-based semi-parametric style transfer framework.
As a whole, the detailed process of training a GNN-based semi-parametric arbitrary stylization model with the proposed algorithm is concluded in Alg.~\ref{alg:A}.

\begin{figure}[!t]
  \centering
  \includegraphics[width=\textwidth]{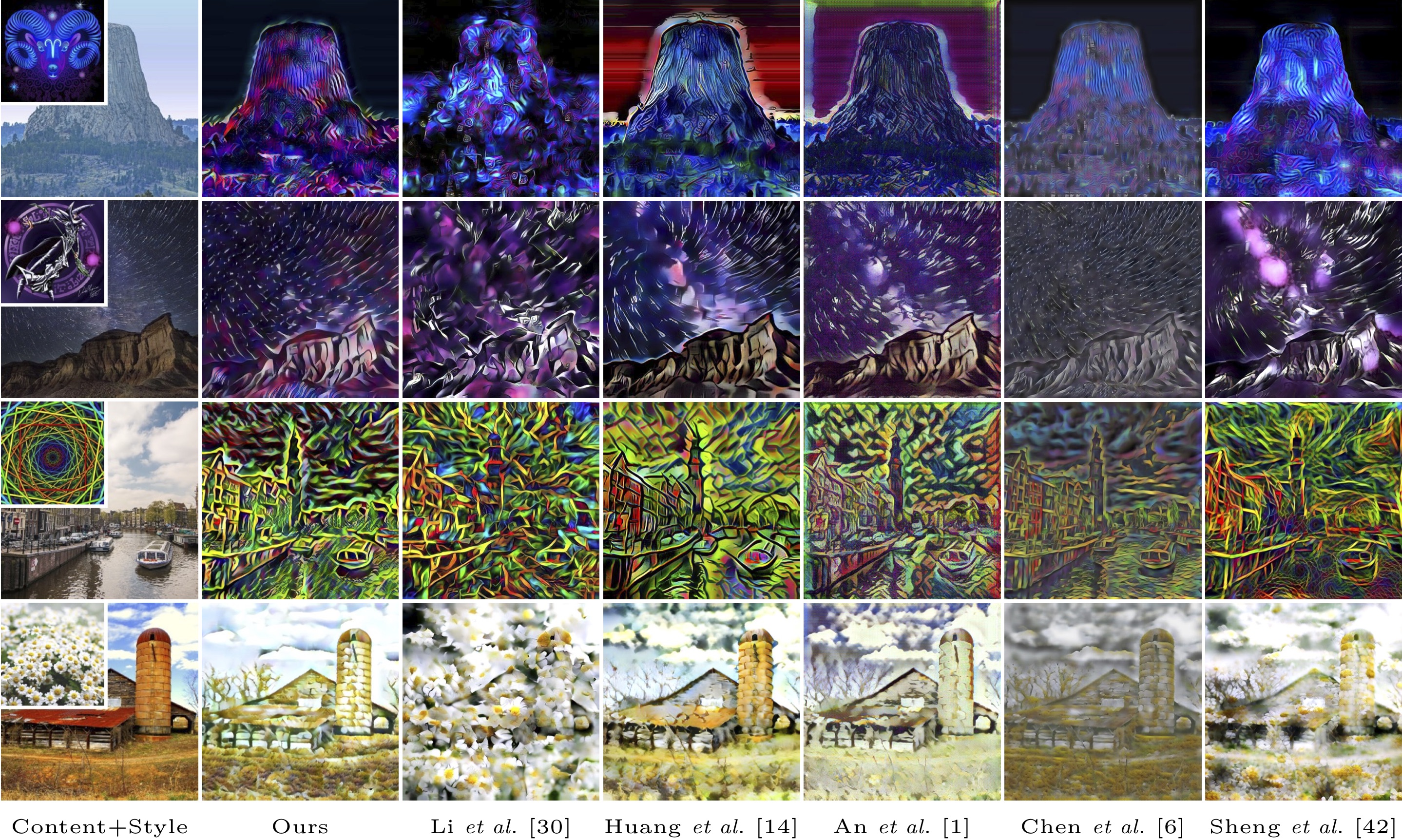}
  % \vspace{-2mm}
  \caption{Qualitative  results of our proposed GNN-based semi-parametric stylization algorithm and other 
  parametric \cite{li2017universal,huang2017arbitrary,an2021artflow} and non-parametric \cite{chen2016fast,sheng2018neural} methods.}
  \label{fig:quality} %% label for entire figure
\end{figure}

% \vspace{-2mm}
\section{Experiments}

% \vspace{-1mm}
\subsection{Experimental Settings}

% \vspace{1mm}
% \noindent\textbf{Implementation Details.} 
We demonstrate here the implementation details as per the stage of the proposed semi-parametric pipeline.
For the stylization graph construction stage, we set $k$ as 5 by default for the NCC-based KNN search.
The stride $s$ for the sliding window is set to 1, whereas the kernel size is set to $5 \times 5$.
At the stage of deformable graph convolution,
we primarily use the graph attention network (GAT) \cite{velivckovic2018graph} for the GNN layers to validate the effectiveness of the proposed semi-parametric NST scheme.
During training, we adopt the Adam optimizer \cite{kingma2014adam} to optimize the whole GNN-based network.
The learning rate is $1 \times 10^{-4}$ with a weight decay of $5 \times 10^{-5}$.
The batch size is set to 8.
The weighting factor $\lambda$ is set to 10.  
We employ a pre-trained VGG-19 as our loss network, as also done in \cite{gatys2016image,huang2017arbitrary}.
The network is trained on the Microsoft COCO dataset \cite{lin2014microsoft} and the WikiArt \cite{wikiart} dataset.
Our code is based on Deep Graph Library (DGL) \cite{wang2019deep}.
The training takes roughly two days on an NVIDIA Tesla A100 GPU.

\subsection{Results}

% \vspace{-1mm}
\noindent\textbf{Qualitative comparison.}
Fig.~\ref{fig:quality} demonstrates the results of the proposed GNN-based semi-parametric method and other arbitrary style transfer methods \cite{li2017universal,huang2017arbitrary,an2021artflow,chen2016fast,sheng2018neural}.
The results of \cite{li2017universal} are prone to distorted patterns.
By contrast, the algorithms of \cite{huang2017arbitrary,an2021artflow} generate sharper details;
however, the local style patterns in their results are not well aligned with the target ones, where very few fine strokes are produced for most styles.
For the non-parametric NST approaches of \cite{chen2016fast,sheng2018neural}, their stylized results either introduce fewer style patterns or suffer from artifacts, due to the potential issue of one-to-one patch mismatching. 
Compared with other approaches, our semi-parametric framework leads to few artifacts, and meanwhile preserves both the global style appearance and the local fine details, thanks to the local patch-based manipulation module with GNNs.

% \vspace{1mm}
\noindent\textbf{Efficiency analysis.} In Tab.~\ref{tab:speed}, we compare the average stylization speed of the proposed approach with other algorithms.
For a fair comparison, all the methods are implemented with PyTorch.
The experiments are performed over 100 equal-size content and style images of different resolutions using an NVIDIA Tesla A100 GPU.
Our speed is, in fact, bottlenecked by the KNN search process, which can be further improved with an optimized KNN implementation.

\begin{table}[!t]
  % \vspace{-6 mm}
  \caption{Average speed comparison in terms of seconds per image. }
  % \vspace{-6mm}
  \begin{center}
  \setlength\tabcolsep{12 pt}
  {\renewcommand{\arraystretch}{1}
  \begin{tabular}{c|cccc}
    \noalign{\hrule height 0.8pt}
    \multirow{2}{*}{\textbf{Methods}} &\multicolumn{3}{c}{\textbf{Time (s)}}\\ \cline{2-4} & \textbf{$256 \times 256$} & \textbf{$384 \times 384$} & \textbf{$512 \times 512$} \\
    \noalign{\hrule height 0.5pt}
      Li \etal \cite{li2017universal} &  0.707 &  0.779 &  0.878 \\
      Huang \etal \cite{huang2017arbitrary} &  0.007 &  0.010 &  0.017  \\
      An \etal \cite{an2021artflow} &  0.069 & 0.108 &  0.169 \\
      Chen \etal \cite{chen2016fast} &  0.017 &  0.051 &  0.218 \\
      Sheng \etal \cite{sheng2018neural} &  0.412 &  0.536 &  0.630 \\
    \hline
      Ours &  0.094 &  0.198 &  0.464 \\
    \noalign{\hrule height 0.8pt}
  \end{tabular}}
  \end{center}
  \label{tab:speed}
  % \vspace{-1.4cm}
  \end{table}

% \vspace{1mm}

% \vspace{-7mm}
\subsection{Ablation Studies}
% \vspace{-1mm}

\begin{figure}[!b]
    \centering
    \includegraphics[width=\textwidth]{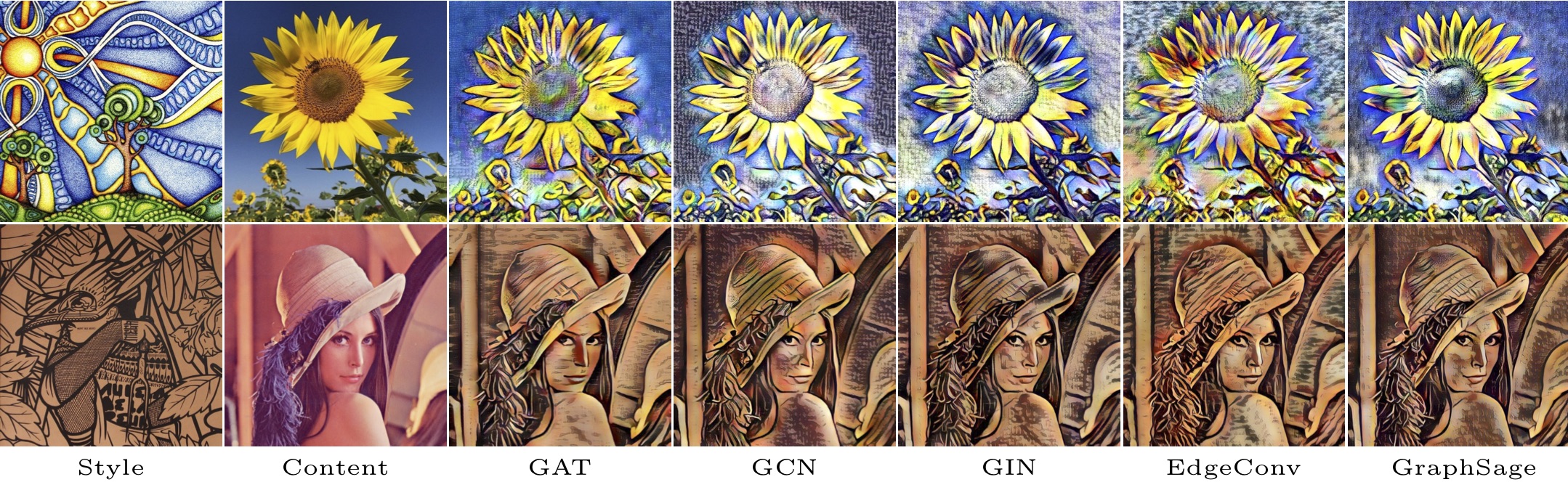}
    % \vspace{-2mm}
    \caption{Comparative results of using various aggregation mechanisms for heterogeneous message passing, including graph attention network (GAT) \cite{velivckovic2018graph}, graph convolutional network (GCN) \cite{kipf2017semi},  graph isomorphism network (GIN) \cite{xu2018powerful}, dynamic graph convolution (EdgeConv) \cite{dgcnn}, and GraphSage \cite{hamilton2017inductive_graphsage}. The GAT mechanism generally yields superior stylization results, thanks to its attention-based aggregation scheme in Eq.~\ref{eq:atten}.}
    \label{fig:gnn} %% label for entire figure
    % \vspace{-0.25cm}
    % \vspace{-0.6cm}
  \end{figure}

\noindent\textbf{Heterogeneous aggregation schemes.}
We show in Fig.~\ref{fig:gnn} the stylization results by using different neighborhood aggregation strategies in the local patch-based manipulation module.
The results of the GAT aggregation scheme, as shown in the \nth{3} column of Fig.~\ref{fig:gnn}, outperform those of others in finer structures and global coherence (the areas of the sky and the human face in Fig.~\ref{fig:gnn}), thereby validating the superiority of the attention scheme in Eq.~\ref{eq:atten}.

\begin{figure}[!t]
    \centering
    \includegraphics[width=\textwidth]{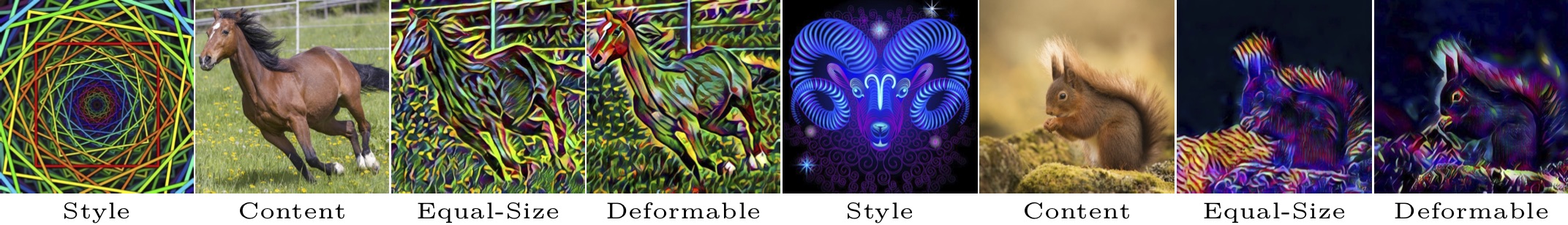}
    % \vspace{-2mm}
    \caption{Results of the equal-size patch division method and the proposed deformable one with a learnable scale predictor. Our deformable scheme allows for cross-scale style-content matching, thereby leading to spatially-adaptive multi-stroke stylization with an enhanced semantic saliency (\eg, the foreground regions of the horse and squirrel).}
    \label{fig:deformable} %% label for entire figure
    % \vspace{-0.6cm}
  \end{figure}

\noindent\textbf{Stylization w/ and w/o the deformable scheme.}
Fig.~\ref{fig:deformable} demonstrates the results with the equal-size patch division method, and those with the proposed deformable patch splitting scheme.
The devised deformable module makes it possible to adaptively control the strokes in different areas.
As a result, the contrast information in the stylized results can be enhanced.

\begin{figure}[!t]
    \centering
    \includegraphics[width=\textwidth]{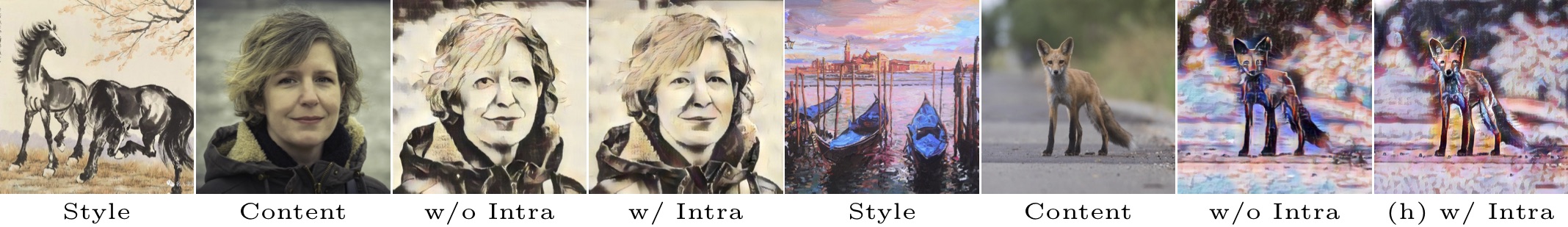}
    % \vspace{-2mm}
    \caption{Results of removing the content-to-content intra-domain edges (w/o Intra) and those with the intra-domain ones (w/ Intra). The devised intra-domain connections incorporate the inter-relationship between the stylized patches at different locations, thereby maintaining the global stylization coherence (\eg, the eye regions in the figure).}
    \label{fig:intra} %% label for entire figure
    % \vspace{-0.6cm}
  \end{figure}

\noindent\textbf{Graph w/ and w/o intra-domain edges.}
In Fig.~\ref{fig:intra}, we validate the effectiveness of the proposed content-to-content message passing scheme, which typically leads to more consistent style patterns in semantically-similar content regions, as can be observed in the foreground human and fox eye areas, as well as the background regions of Fig.~\ref{fig:intra}.

\begin{figure}[!b]
  \centering
  \includegraphics[width=\textwidth]{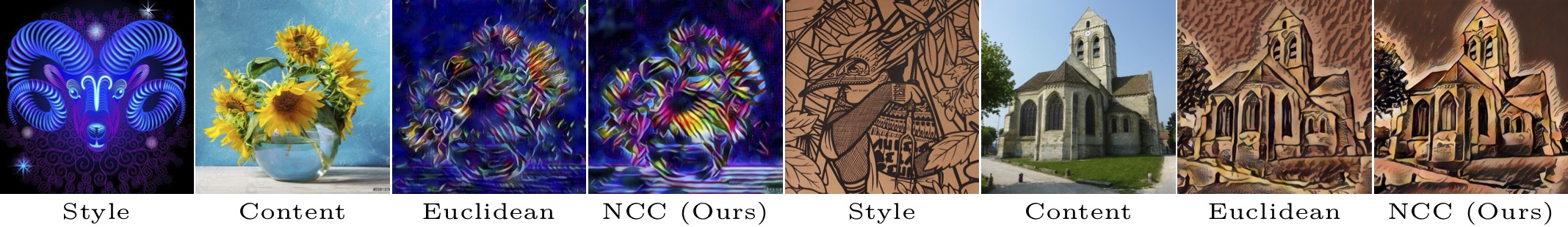}
  % \vspace{-2mm}
  \caption{Results obtained using Euclidean distance and normalized cross-correlation (NCC) for similarity measurement during the construction of heterogeneous edges.}
  % content-to-style and content-to-content edges.}
  \label{fig:distance} %% label for entire figure
  % \vspace{-0.6cm}
\end{figure}

\noindent\textbf{Euclidean distance \emph{vs.} normalized cross-correlation.}
Fig.~\ref{fig:distance} shows the results of using the Euclidean distance and the normalized cross-correlation (NCC) as the distance metric, respectively, in the construction of the stylization graph.
The adopted metric of NCC in our framework, as observed from the \nth{4} and \nth{8} columns of Fig.~\ref{fig:distance}, leads to superior 
performance than the Euclidean distance (Fig.~\ref{fig:distance}, the \nth{3} and \nth{7} columns) in terms of both the global stroke arrangements and local details.

\begin{figure}[!t]
  \centering
  \includegraphics[width=\textwidth]{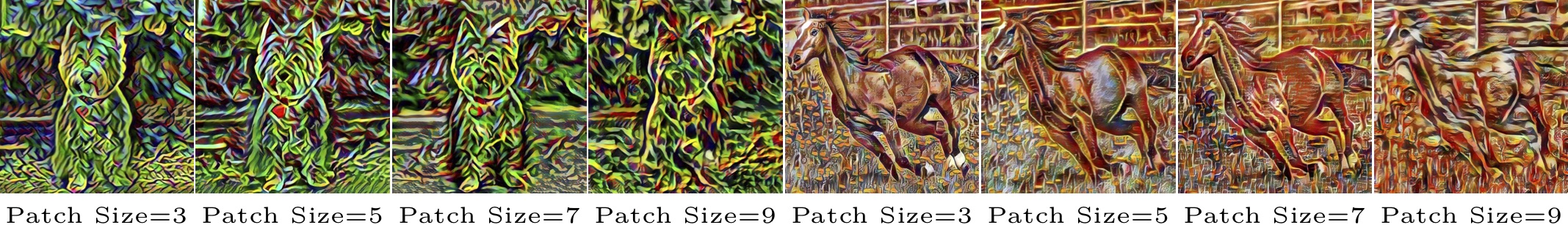}
  % \vspace{-2mm}
  \caption{Results obtained using various patch sizes for constructing content and style vertices in the local patch-based manipulation module. By using a larger patch size, the stylized results can maintain an overall larger stroke size.}
  \label{fig:patchsize} %% label for entire figure
  % \vspace{-0.4cm}
\end{figure}

\noindent\textbf{Various patch sizes.}
We show in Fig.~\ref{fig:patchsize} the results of diversified feature patch sizes. Larger patch sizes, as shown from the left to right in the figure, generally lead to larger strokes in the stylized results, which is especially obvious when we observe the regions of the dog and horse in Fig.~\ref{fig:patchsize}.

\subsection{Diversified Stylization Control}

\begin{figure}[!t]
    \centering
    \includegraphics[width=\textwidth]{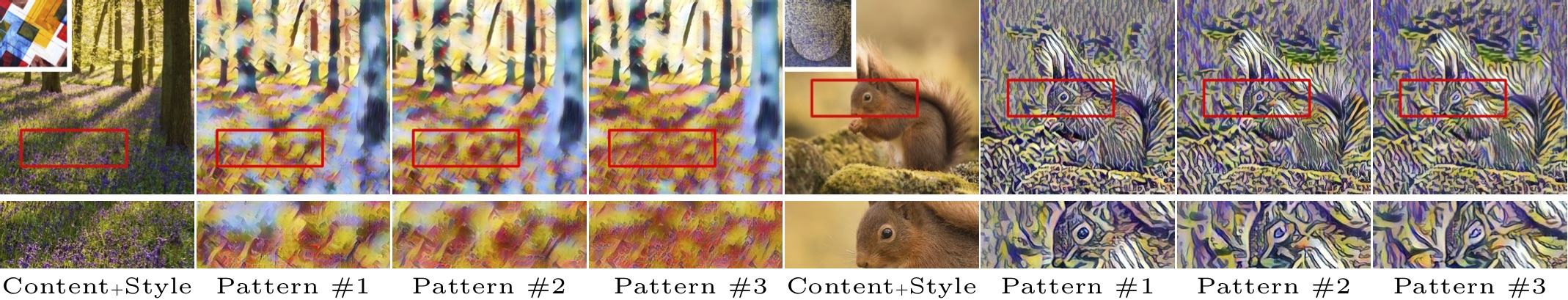}
    % \vspace{-2mm}
    \caption{Flexible control of diversified patch-based arbitrary style transfer during inference. The proposed GNN-based semi-parametric stylization scheme makes it possible to generate heterogeneous style patterns with only a single trained model. }
    % We also zoom in on the same regions (\ie, the red frames in Fig.~\ref{fig:diverse}) to observe the details.}
    \label{fig:diverse} %% label for entire figure
    % \vspace{-0.4cm}
    \end{figure}

The proposed GNN-based arbitrary style transfer scheme, as shown in Fig.~\ref{fig:diverse}, can readily support diversified stylization with solely a single model.
We also zoom in on the same regions (\ie, the red frames in Fig.~\ref{fig:diverse}) to observe the details.
Such diversities in Fig.~\ref{fig:diverse} are specifically achieved by simply changing the numbers of node-specific connections for heterogeneous message passing, which provide users of various tastes with more stylization choices.

% \vspace{-0.5mm}
\section{Conclusions}
% \vspace{-1mm}

In this paper, we introduce a semi-parametric arbitrary style transfer scheme for the effective transfers of challenging style patterns at the both local and global levels.
Towards this goal, we identify two key challenges in existing parametric and non-parametric stylization approaches, and propose a dedicated GNN-based style transfer scheme to solve the dilemma.
This is specifically accomplished by modeling the style transfers as the heterogeneous information propagation process among the constructed content and style vertices for accurate patch-based style-content correspondences.
Moreover, we develop a deformable graph convolutional network for various-scale stroke generations. 
Experiments demonstrate that the proposed approach achieves favorable performance in both global stroke arrangement and local details.
In our future work, we will strive to generalize the proposed GNN-based scheme to other vision tasks.

\paragraph{\bf Acknowledgments.}
% This work is supported by ARC FL-170100117, 
% AI Singapore (Award No.: AISG2-RP-2021-023), 
% and NUS Faculty Research Committee Grant (WBS: A-0009440-00-00).
This research is supported by ARC FL-170100117, the National Research Foundation Singapore under its AI Singapore Programme (Award Number: AISG2-RP-2021-023), and NUS Faculty Research Committee Grant (WBS: A-0009440-00-00).

% Mr Yongcheng Jing is supported by ARC FL-170100117. Dr Xinchao Wang is supported by AI Singapore (Award No.: AISG2-RP-2021-023) and NUS Faculty Research Committee Grant (WBS: A-0009440-00-00).

%
\bibliographystyle{splncs04}
\bibliography{MYRE}
\end{document}